\documentclass{article}
\usepackage{spconf,amsmath,graphicx}

\usepackage{enumitem}
\setlist{nosep, leftmargin=14pt}

\usepackage{mwe} 

\usepackage{makecell}
\usepackage{amssymb}

\usepackage{booktabs} 
\usepackage{makecell} 

\title{RDTE-UNet: A Boundary and Detail Aware UNet for Precise Medical Image Segmentation}
\twoauthors
 {Jierui Qu}
	{National University of Singapore\\
	College of Design and Engineering\\
	117575, Singapore}
 {Jianchun Zhao\sthanks{Corresponding Author, email: zhao\_jianchun@stu.xjtu.edu.cn.}}
	{Xi'an Jiaotong University\\
	School of Electrical Engineering\\
	Xi'an 710049, China}
\begin{document}
\ninept
\maketitle
\begin{abstract}
Medical image segmentation is essential for computer‑assisted diagnosis and treatment planning, yet substantial anatomical variability and boundary ambiguity hinder reliable delineation of fine structures. We propose RDTE‑UNet, a segmentation network that unifies local modeling with global context to strengthen boundary delineation and detail preservation. RDTE‑UNet employs a hybrid ResBlock detail‑aware Transformer backbone and three modules: ASBE for adaptive boundary enhancement, HVDA for fine‑grained feature modeling, and EulerFF for fusion weighting guided by Euler’s formula. Together, these components improve structural consistency and boundary accuracy across morphology, orientation, and scale. On Synapse and BUSI dataset, RDTE-UNet has achieved a comparable level in terms of segmentation accuracy and boundary quality.\footnote{Available after paper is accepted.}
\end{abstract}
\section{Introduction}
Medical image segmentation is a core task in medical image analysis, partitioning complex scans into anatomically meaningful regions and enabling precise extraction of organs and lesions for diagnosis and treatment planning. However, manual delineation by experts is time-consuming, subjective, and error-prone~\cite{zhou_bsbp-rwkv_2024}, underscoring the need for automated and accurate methods to streamline clinical workflows.

Computer-aided medical image analysis is pivotal to modern healthcare. Deep learning (DL) techniques~\cite{lecun_deep_2015}, which learn complex patterns directly from imaging data, have advanced segmentation and improved accuracy and efficiency. Among these methods, U-Net~\cite{ronneberger_u-net_2015} is widely adopted for its U-shaped, symmetric encoder–decoder design: the encoder captures high-level semantics via downsampling, while the decoder combines upsampling with skip connections to recover fine details, yielding strong performance in medical image segmentation.

The success of the U-Net architecture has led to the development of numerous variants, which primarily enhance the original network using Convolutional Neural Network (CNN)~\cite{huang_unet_2020,wu_multi-scale_2024,zhu_selfreg-unet_2024} or Transformer~\cite{cao_swin-unet_2023,huang_missformer_2021}. CNN are widely adopted for their strong capability in extracting local features, but the intrinsic locality of convolutional operations limits their ability to capture long-range dependencies~\cite{chen_transunet_2021}. In contrast, the Transformer architecture excels at modeling long-range dependencies but tends to be less effective in extracting fine-grained local features~\cite{chen_bootstrapping_2024,chen_mim-istd_2024}.

To address the respective limitations of CNNs and Transformers, hybrid models integrate both architectures to couple CNNs’ strong local feature extraction with Transformers’ long-range dependency modeling~\cite{chen_ms-unet_2024,chen_transunet_2021,wang_mixed_2022,xu_levit-unet_2024}. TransUNet~\cite{chen_transunet_2021} augments U-Net with Transformer-based global context to improve segmentation, while Wang et al.~\cite{wang_mixed_2022} propose a mixed Transformer module (MTM) that jointly learns intra- and inter-sample correlations via Local–Global Gaussian-weighted Self-Attention (LGG-SA) and External Attention (EA). Despite state-of-the-art results on specific tasks, these hybrids still struggle with targets exhibiting large variations in orientation, shape, and scale, and remain limited in capturing fine-grained details such as boundaries and microstructures.

To address these aforementioned challenges, we propose RDTE-UNet, a ResBlock–Details Transformer–based segmentation network that strengthens boundary and detail delineation, mitigating boundary blur and fine-structure loss. RDTE-UNet comprises an Adaptive Shape-aware Boundary Enhancement (ASBE), a Horizontal–Vertical Detail Attention (HVDA), and an Euler Feature Fusion (EulerFF) module. ASBE first extracts initial features; its ARConv~\cite{wang_adaptive_2025} adaptively adjusts kernel sizes and sampling locations to organ/lesion morphology, enabling multi-scale representation and differential boundary enhancement. A subsequent ResBlock deepens local features, while features are concurrently routed to a Details Transformer for global, context-aware detail modeling. Within it, HVDA emphasizes subtle structures along horizontal and vertical directions, improving recognition of fine details and complex topologies. During decoding, EulerFF fuses multi-scale encoder–decoder features via an Eulerian weighting that dynamically modulates horizontal, vertical, and channel dimensions to prioritize critical boundaries and details. This design yields more complete and accurate segmentation of targets with complex topology.

The main contributions of this study can be summarized as follows:
\begin{enumerate}
    \item We introduce ASBE, which dynamically adapts convolutional kernels to target morphology to extract multi-scale cues and sharpen boundary details.
    \item We design HVDA to strengthen the Transformer’s fine-grained modeling via a StairConv with a tailored receptive field.
    \item We propose EulerFF, which employs Eulerian weighting to dynamically modulate and efficiently fuse multi-scale encoder–decoder features, enhancing anisotropic detail perception and segmentation completeness under complex topologies.
    \item We conduct extensive experiments on Synapse~\cite{landman_miccai_2015} and BUSI~\cite{al-dhabyani_dataset_2020}, where RDTE-UNet surpasses state-of-the-art methods in accuracy and detail preservation.
\end{enumerate}

\begin{figure*}[htb]
\centering
\includegraphics[width=0.99\textwidth]{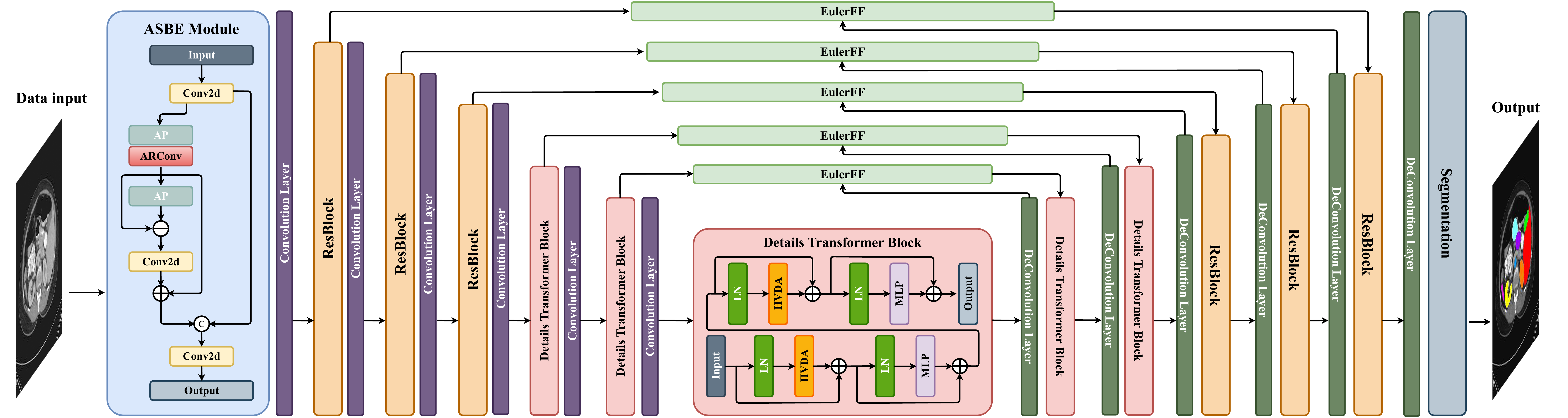}
\caption{Overview of the framework.} \label{fig:all}
\end{figure*}

\section{Methods}
RDTE-UNet (Fig.~\ref{fig:all}) comprises ASBE, ResBlocks, Details Transformer blocks, and EulerFF. For an input image of size $H\times W\times C$, ASBE performs initial feature extraction and boundary enhancement. The encoder has five stages: the first three use a standard residual block~\cite{he_deep_2016} , and the last two adopt the Details Transformer. Each encoder stage ends with a $2\times2$ stride-2 convolution that halves spatial resolution while doubling channels. The decoder mirrors the encoder with five stages; a deconvolution layer doubles spatial resolution and halves channels at each stage. EulerFF operates on the skip connections to fuse shallow, high-resolution encoder features with deep, semantic decoder features, mitigating information loss from downsampling. Module architectures and functions are detailed in the following sections.


\subsection{Adaptive Shape-aware Boundary Enhancement Module (ASBE)}
For input images, ASBE performs initial feature extraction and boundary enhancement. It integrates an Adaptive Rectangular Convolution (ARConv) that dynamically adjusts kernel size and sampling pattern to target geometry, enabling flexible multi-scale, anisotropic feature capture and addressing target diversity. To further sharpen boundaries, a difference algorithm~\cite{gao_multi-scale_2024} accentuates edge responses in the feature maps. As shown on the left of Fig.~\ref{fig:all}, ASBE first applies a $1\times1$ convolution for channel compression, then extracts shape-aware features via average pooling (AP) and the adaptive ARConv. The difference between pooled and original features is computed and fused with the residual path through a nonlinearity to strengthen boundary cues. Finally, the refined boundary features are concatenated with the compressed features and passed through another $1\times1$ convolution to produce the enhanced feature map. The computation is formulated as follows:

\subsection{Details Transformer Block}
Unlike conventional Transformers that emphasize global context modeling~\cite{vaswani_attention_2017}, the proposed Details Transformer Block is tailored to enhance fine-grained feature representations for medical image segmentation. As shown on the middle of Fig.~\ref{fig:all}, Details Transformer Block stacks two identical submodules, each comprising Layer Normalization (LN), the HVDA module, and a two-layer MLP for nonlinear mapping, with residual connections to stabilize training and preserve information. The HVDA module focuses on enhancing detailed information such as boundaries and microstructures, which are often overlooked by standard self-attention mechanisms.

\begin{figure}
\centering
\includegraphics[width=0.8\columnwidth]{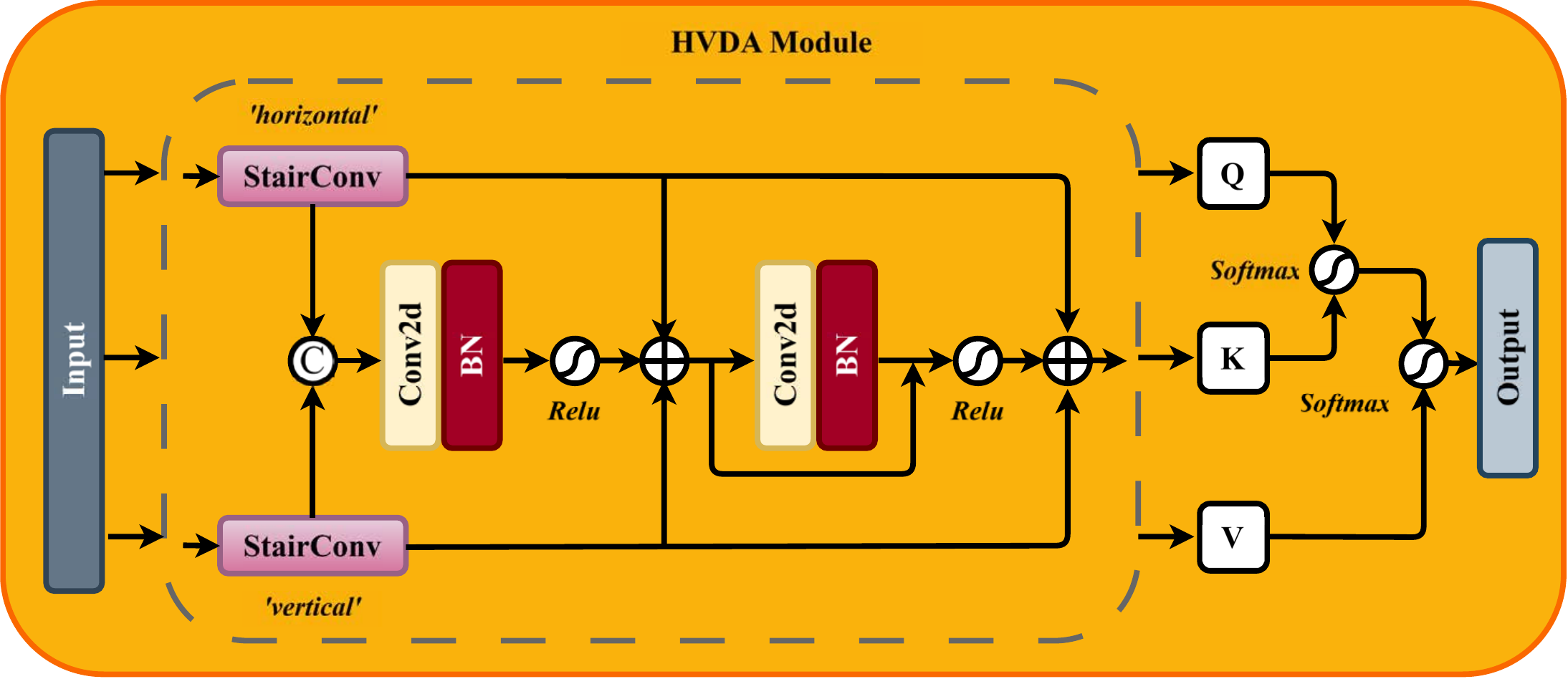}
\caption{Schematic of HVDA Module.} \label{fig:hvda}
\end{figure}

\subsubsection{Horizontal and Vertical Details Self-Attention module (HVDA)}
To better capture fine structures and complex topologies in medical images, we propose HVDA, inspired by Global Spatial Attention (GSA)~\cite{chen_transattunet_2024} (Fig.~\ref{fig:hvda}). Unlike uniform global self-attention, HVDA employs StairConv along horizontal and vertical axes to amplify boundary details and small-scale targets. The extracted features are fused via residual connections and subsequently fed into self-attention for global modeling.

Given the input feature $x_{in}$, HVDA extracts horizontal and vertical detail features $x_{hd}$ and $x_{vd}$ via two parallel StairConv paths and concatenates them. A $1\times1$ convolution reduces channel dimensionality to lower computational cost. To preserve information, the concatenated features undergo residual cascading to yield $x_C$. A $3\times3$ convolutional block then extracts deeper features, with residual concatenation applied before ReLU. Finally, the horizontal-vertical detail features are added to the deep features to produce the fused representation $x_{fusion}$. Multiple residual operations mitigate feature information loss. The corresponding equations are as follows:

\begin{equation}
x_{hd}=\mathrm{StairConv}_h(x_{in})
\end{equation}
\begin{equation}
x_{vd}=\mathrm{StairConv}_v(x_{in})
\end{equation}
\begin{equation}
x_{cat}=\mathrm{ReLU}\left(\mathrm{BN}\left(\mathrm{Conv}2\mathrm{d}\left(\mathrm{Concat}(x_{hd},x_{vd})\right)\right)\right)+x_{hd}+x_{vd}
\end{equation}
\begin{equation}
x_{fusion}=\mathrm{ReLU}\left(\mathrm{BN}\left(\mathrm{Conv}2\mathrm{d}(x_C)\right)+x_C\right)+x_{hd}+x_{vd}
\end{equation}

The HVDA module extracts the detail features and fuses the input features through three identical above structures to obtain the corresponding $x_{fusion}$, and then projects them to the three embedding spaces to obtain the Query $Q\in\mathbb{R}^{hw}$, Key $K\in\mathbb{R}^{hw}$, and Value $V\in\mathbb{R}^{hw}$, respectively. Subsequently, matrix multiplication is performed on $Q$ and $K$ and subjected to Softmax normalization to obtain the attention feature map $B$, which is computed as follows:

\begin{equation}
B=\frac{\exp{(Q\cdot K)}}{\sum_{n=1}^{hw}\exp{(Q\cdot K)}}
\end{equation}

The final feature representation is obtained by weighting the value vector $V$ using the attention feature map $B$. The formula is as follows:

\begin{equation}
\mathrm{HVDA}(x_{in})=V\cdot B
\end{equation}

\begin{figure}
\centering
\includegraphics[width=0.8\columnwidth]{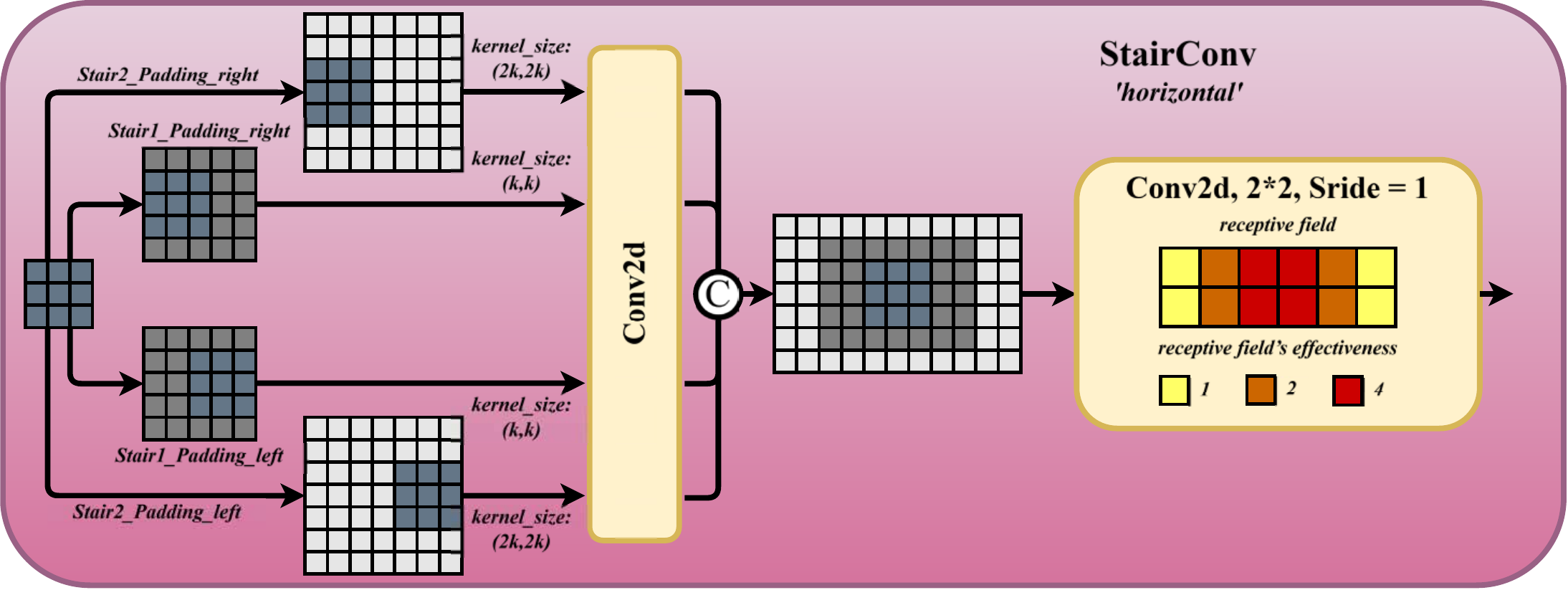}
\caption{Schematic of StairConv.} \label{fig:stairconv}
\end{figure}

\subsubsection{StairConv}
We propose StairConv, a convolution module that progressively enlarges the receptive field via multi-scale, stepwise asymmetric padding and convolution to capture fine-grained details (e.g., boundaries and microstructures) within feature maps. As shown in Fig.~\ref{fig:stairconv}, StairConv adopts stepwise offset padding and is instantiated in horizontal and vertical variants. Let the input tensor be $x_{\mathrm{in}}\in\mathbb{R}^{h_{0}\times w_{0}\times c_{in}}$, where $h_0$, $w_0$, and $c_{in}$ denote height, width, and channel count, respectively. StairConv comprises two levels of offset convolutional branches at different scales; each level includes right (or upward) and left (or downward) shift branches. For example, horizontal StairConv is computed as follows:

\begin{equation}
F_{1}^{right/left}=\mathrm{SiLU}\left(\mathrm{BN}\left(\mathrm{Conv}_{k,k}\left(\mathcal{P}_{1}^{right/left}(x_{in})\right)\right)\right)
\end{equation}
\begin{equation}
F_2^{right/left}=\mathrm{SiLU}\left(\mathrm{BN}\left(\mathrm{Conv}_{2k,2k}\left(\mathcal{P}_2^{right/left}(x_{in})\right)\right)\right)
\end{equation}
where $\mathcal{P}_i^{side}(\cdot)$ denotes a predefined asymmetric padding operation on one side, and $\mathrm{Conv}_{m,n}(\cdot)$ represents an $m\times n$ convolutional operation without padding. Subsequently, the four intermediate features are concatenated along the channel dimension:

\begin{equation}
F_{cat}=\mathrm{Concat}\left(F_1^{right},F_1^{left},F_2^{right},F_2^{left}\right)
\end{equation}

A tensor of size $h_1\times w_1\times4c^{\prime}$ is obtained, where $c^{\prime}$ denotes the number of channels output from each branch.

Finally, the concatenated features are integrated using a $2\times2$ convolution without padding:

\begin{equation}
F_{out}=\mathrm{SiLU}\left(\mathrm{BN}\left(\mathrm{Conv}_{2,2}(F_{cat})\right)\right)
\end{equation}

The final output $F_{out}\in\mathbb{R}^{h_2\times w_2\times c_{out}}$ is obtained, where $c_{out}$ is the number of target output channels.

As shown in Fig.~\ref{fig:stairconv}, StairConv achieves a wider and denser receptive field than traditional convolution by employing a multi-scale stacking design. Additionally, the receptive fields at different spatial locations exhibit varying response strengths, which contributes to enhanced representation of fine image details.

\begin{figure}[t]
\centering
\includegraphics[width=0.8\columnwidth]{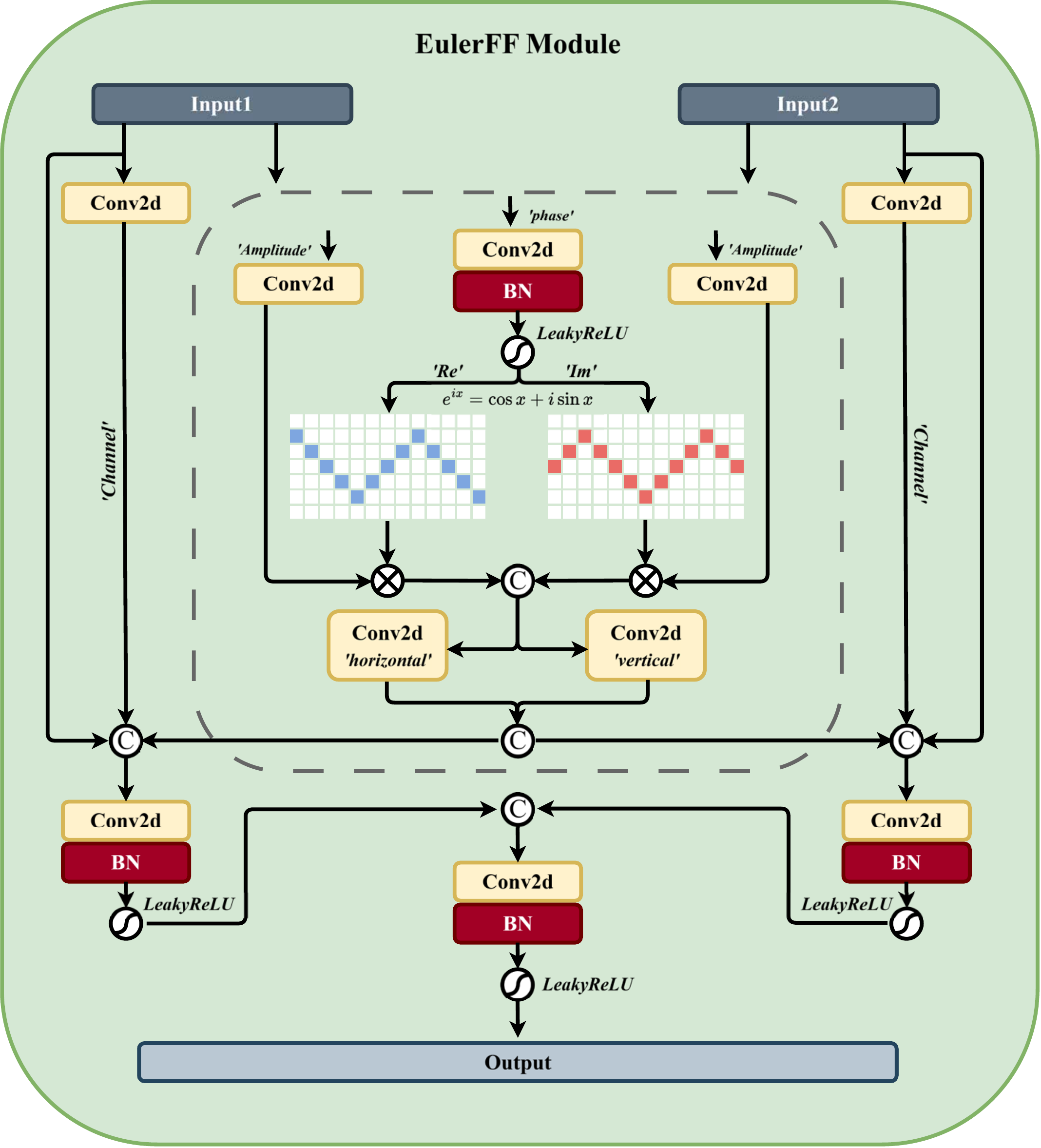}
\caption{Schematic of Euler Module.} \label{fig:eulerff}
\end{figure}

\begin{table*}
\centering
\caption{Experimental results of the Synapse Dataset. DSC of each single class is also presented.}\label{table:Synapse}
\begin{tabular}{ccccccccccc}
\hline
Method & \makecell{DSC(\%)$\uparrow$} & \makecell{HD95(mm)$\downarrow$} & Aorta & \makecell{Gall.} & \makecell{Kid(L)} & \makecell{Kid(R)} & Liver & \makecell{Panc.} & Spleen & Stomach\\
\hline
Trans-UNet & 79.15 & 28.47 & 87.95 & 67.08 & 80.58 & 78.87 & 93.92 & 61.17 & 86.52 & 77.12\\
Swin-UNet & 81.03 & 19.54 & 86.84 & 70.40 & 83.76 & 80.13 & 93.57 & 67.87 & 90.23 & 78.43\\
MT-UNet & 80.72 & 22.48 & 87.56 & 70.86 & 83.51 & 79.15 & 92.89 & 67.83 & 87.60 & 76.33\\
RWKV-UNet & 85.62 & 14.83 & {\bfseries 89.98} & 72.89 & 88.27 & 85.24 & 95.06 & 77.69 & 90.15 & {\bfseries 85.76}\\
{\bfseries Ours} & {\bfseries 86.63} & {\bfseries 11.69} & 89.86 & {\bfseries 81.96} & {\bfseries 90.06} & {\bfseries 89.44} & 87.24 & {\bfseries 83.07} & {\bfseries 91.24} & 80.19\\
\hline
\end{tabular}
\end{table*}

\begin{table}
\centering
\caption{Experimental results of the BUSI Dataset.}\label{table:BUSI}
\begin{tabular}{ccc}
\hline
Method & DSC(\%)$\uparrow$ & HD95(mm)$\downarrow$\\
\hline
Trans-UNet & 60.42 & 32.78\\
Swin-UNet & 62.91 & 30.67\\
MT-UNet & 62.13 & 39.08\\
RWKV-UNet & 64.85 & 29.57\\
{\bfseries Ours} & {\bfseries 66.31} & {\bfseries 27.73}\\
\hline
\end{tabular}
\end{table}

\subsection{Euler feature fusion Module (EulerFF)}

To further strengthen encoder–decoder multi-scale interaction and improve perception and integration of complex topologies and anisotropic details, we introduce EulerFF, a feature fusion module grounded in Euler’s formula (Fig.~\ref{fig:eulerff}). EulerFF constructs joint representations along horizontal, vertical, and channel dimensions by dynamically modulating feature amplitude and phase, enabling efficient multi-scale fusion. Inspired by Eulerian weighting~\cite{tian_eulernet_2023}, it models features in complex form to heighten sensitivity to directional details.

In this module, features are represented as complex-valued expressions composed of magnitude–phase pairs, and the following transformations are applied:

\begin{equation}
\mathcal{F}_{Euler}=A\cdot\cos\left(\theta\right)+j\cdot A\cdot\sin\left(\theta\right)
\end{equation}
where $A$ denotes the feature amplitude and $\theta$ denotes the direction-sensitive phase learned by the phase modulator. For the input features, the horizontal and vertical submodules yield eigen-amplitudes $A_h$, $A_v$ and eigen-phases $\theta_h$, $\theta_v$, which are expanded into Euler-based feature representations by concatenating their real and imaginary components:

\begin{equation}
\mathcal{F}_{h/v}=\mathrm{Concat}(A_{h/v}\cdot\cos{(\theta_{h/v})},A_{h/v}\cdot\sin{(\theta_{h/v})})
\end{equation}

Subsequently, grouped convolution performs directional modeling on the horizontal feature $\mathcal{F}_{h}$ and vertical feature $\mathcal{F}_{v}$ to produce anisotropic response–enhanced features $\mathcal{T}_{h}$ and $\mathcal{T}_{v}$, while channel-wise extraction on the input yields $\mathcal{T}_{c}$. The original and directionally enhanced features are then concatenated, and the combined tensor is passed through a dimension-reducing fusion layer to integrate information:

\begin{equation}
\mathrm{FusionLayer}(x_{in})=\mathrm{Concat}(x_{in},\mathcal{T}_h,\mathcal{T}_v,\mathcal{T}_c)
\end{equation}
Let $x_{s}$ and $x_{d}$ be the input features to the Euler module from the skip connections and decoder , resulting in the outputs $\mathcal{F}_{s}$ and $\mathcal{F}_{d}$.

The outputs of the two processing streams are further concatenated and fused into a final output feature:

\begin{equation}
\mathcal{F}_{out}=\mathrm{FusionLayer}(\mathrm{Concat}(\mathcal{F}_s,\mathcal{F}_d))
\end{equation}

\section{Experiments}
\subsection{Datasets and Metrics}
We evaluate on the Synapse Multi-organ Segmentation dataset (Synapse)~\cite{landman_miccai_2015} and the Breast Ultrasound Image dataset (BUSI)~\cite{al-dhabyani_dataset_2020}. Synapse contains 3,779 abdominal axial CT images; we use a 60/40 train–test split to segment eight organs (aorta, gallbladder, spleen, left kidney, right kidney, liver, pancreas, and stomach). BUSI comprises 780 ultrasound images labeled benign (56.0\%), malignant (26.9\%), or normal (17.1\%)~\cite{aumente-maestro_multi-task_2025}; we adopt a 70/30 split and include all categories. Following~\cite{jiang_rwkv-unet_2025,xu_levit-unet_2024}, evaluation uses Dice Similarity Coefficient (DSC) and 95\% Hausdorff Distance (HD95).

\begin{table}[t]
\centering
\caption{Ablation study on Synapse dataset.}\label{table:Ablation}
\begin{tabular}{ccc}
\hline
Method & DSC(\%)$\uparrow$ & HD95(mm)$\downarrow$\\
\hline
{Ours w/o ARBE} & 84.97 & 15.17\\
{Ours w/o HVDA} & 82.76 & 14.83\\
{Ours w/o EulerFF} & 80.98 & 17.49\\
Ours & {\bfseries 86.63} & {\bfseries 11.69}\\
\hline
\end{tabular}
\end{table}

\begin{figure}[t]
\includegraphics[width=\columnwidth]{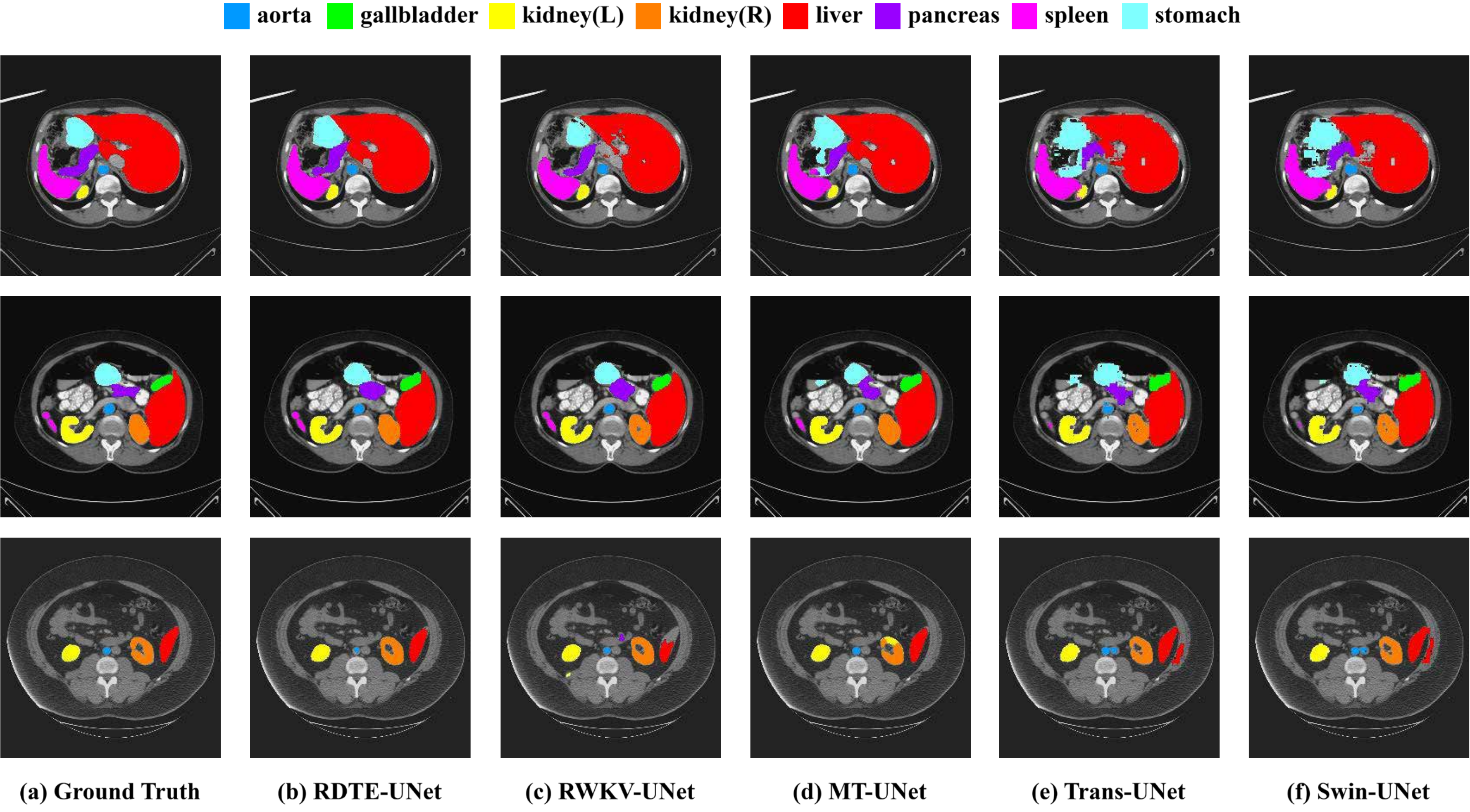}
\caption{Qualitative comparison of different methods through visualization on Synapse dataset. Our method produces fewer false positives and better preserves fine details.} \label{fig:visual}
\end{figure}

\subsection{Experimental results and visualization}
Table~\ref{table:Synapse} compares RDTE-UNet with some traditional methods on Synapse. RDTE-UNet achieves the best results—86.63\% (DSC$\uparrow$) and 11.69 mm (HD95$\downarrow$); the HD95 gain indicates more accurate boundary localization. Qualitative results in Fig.~\ref{fig:visual} further show clear advantages in capturing fine structures, boundary details, and complex topologies. On BUSI (Table~\ref{table:BUSI}), RDTE-UNet attains 66.31\% DSC and 27.73 mm HD95, demonstrating robustness and cross-modality generalization.

\subsection{Ablation Study}
In order to evaluate the effectiveness of each proposed module, we conducted ablation experiments on the Synapse dataset, as summarized in Table~\ref{table:Ablation}. Specifically, we first removed the ASBE module, resulting in a decrease in DSC to 84.97\% and an increase in HD95 to 15.17 mm. Next, we replaced the proposed HVDA module with the simpler GSA module~\cite{chen_transattunet_2024} , and also tested the removal of the EulerFF module, where the encoder and decoder were connected using a standard skip connection instead. The experimental results show that using the HVDA module increases the DSC and HD95 by 3.87\% and 3.14 mm, respectively, while using the EulerFF module significantly enhances the model performance, with an increase of 5.65\% in the DSC and 5.80 mm in the HD95. Generally, the RDTE-UNet outperforms all types of variants in the experiments, suggesting that all the three modules proposed in this study contribute to the model performance improvement.

\section{Conclusion}
In this paper, we propose a novel medical image segmentation network, RDTE-UNet, designed to enhance segmentation performance, particularly in boundary regions and fine structural details. The network adopts a hybrid architecture composed of ResBlock and Details Transformer Block, and incorporates three innovative modules—ASBE, HVDA, and EulerFF—which effectively integrate local feature extraction and global context modeling. This design is optimized for segmentation tasks involving significant morphological variation and complex anatomical structures. Experimental results on the Synapse and BUSI datasets demonstrate that RDTE-UNet surpasses existing state-of-the-art methods in segmentation accuracy, especially in identifying structures with complex topological and morphological characteristics. We believe this study provides a valuable contribution to computer-aided diagnosis and has the potential to assist clinicians in making more accurate and efficient decisions.

\section{ACKNOWLEDGMENTS}
No funding was received for conducting this study. The authors have no relevant financial or non-financial interests to disclose.
\bibliographystyle{IEEEbib}
\bibliography{refs}

\end{document}